\documentclass{article}

\usepackage{PRIMEarxiv}

\usepackage[utf8]{inputenc} 
\usepackage[T1]{fontenc}    
\usepackage{hyperref}       
\usepackage{url}            
\usepackage{booktabs}       
\usepackage{amsfonts}       
\usepackage{nicefrac}       
\usepackage{microtype}      
\usepackage{lipsum}
\usepackage{fancyhdr}       
\usepackage{graphicx}       
\usepackage{cite}
\usepackage{amsmath,amssymb,amsfonts}
\usepackage{algorithmic}
\usepackage{graphicx}
\usepackage{textcomp}
\usepackage{booktabs}
\usepackage{diagbox}
\usepackage[utf8]{inputenc} 
\usepackage[T1]{fontenc}    
\usepackage{hyperref}       
\usepackage{url}            
\usepackage{booktabs}       
\usepackage{amsfonts}       
\usepackage{nicefrac}       
\usepackage{microtype}      
\usepackage{lipsum}		
\usepackage{graphicx}
\usepackage{doi}
\usepackage[justification=centering]{caption}
\graphicspath{{media/}}     

\title{Using frequency attention to make adversarial patch powerful against person detector
\thanks{
The first author and the second author have the same contribution to the work. 
} 
}

\title{STDC-MA Network for Semantic Segmentation}

\author{
 Xiaochun Lei \\
  School of Computer Science and Information Security\\
  Guilin University of Electronic Technology\\
  China, GuiLin 541010 \\
  \texttt{lxc8125@guet.edu.cn} \\
   \And
   Linjun Lu \\
  School of Computer Science and Information Security\\
  Guilin University of Electronic Technology\\
  China, GuiLin 541010 \\
  \texttt{linjunlu@zerorains.top} \\
  \And
  Zetao Jiang$^*$\\
  School of Computer Science and Information Security\\
  Guilin University of Electronic Technology\\
  China, GuiLin 541010 \\
  \texttt{zetaojiang@guet.edu.cn} \\
  \And
  Zhaoting Gong \\
  School of Computer Science and Information Security\\
  Guilin University of Electronic Technology\\
  China, GuiLin 541010 \\
  \texttt{gavin@gong.host} \\
  \And
   Chang Lu \\
  School of Computer Science and Information Security\\
  Guilin University of Electronic Technology\\
  China, GuiLin 541010 \\
  \texttt{Changlu@keter.top} \\
  \And
  Jiaming Liang \\
  School of Computer Science and Information Security\\
  Guilin University of Electronic Technology\\
  China, GuiLin 541010 \\
  \texttt{me@puqing.work} \\
}

\begin{document}
\maketitle

\begin{abstract}
	Semantic segmentation is applied extensively in autonomous driving and intelligent transportation with methods that highly demand spatial and semantic information. Here, an STDC-MA network is proposed to meet these demands. First, the STDC-Seg structure is employed in STDC-MA to ensure a lightweight and efficient structure. Subsequently, the feature alignment module (FAM) is applied to understand the offset between high-level and low-level features, solving the problem of pixel offset related to upsampling on the high-level feature map. Our approach implements the effective fusion between high-level features and low-level features. A hierarchical multiscale attention mechanism is adopted to reveal the relationship among attention regions from two different input sizes of one image. Through this relationship, regions receiving much attention are integrated into the segmentation results, thereby reducing the unfocused regions of the input image and improving the effective utilization of multiscale features. STDC- MA maintains the segmentation speed as an STDC-Seg network while improving the segmentation accuracy of small objects. STDC-MA was verified on the verification set of Cityscapes. The segmentation result of STDC-MA attained 76.81\% mIOU with the input of 0.5x scale, 3.61\% higher than STDC-Seg.
\end{abstract}


\section{Introduction}
Semantic segmentation is s a classic computer vision task adopted widely in autonomous driving, video surveillance, robot perception, etc. Unlike object detection, semantic segmentation aims to achieve pixel-wise classification. It can provide detailed contour and category information of an object when applied in critical fields, including autonomous driving. The analysis of traffic behavior in Smart City and intelligent transportation can become more rational with semantic information. Although semantic segmentation methods are highly developed, much improvement is needed to achieve speed and accuracy in the realistic demand.

The goal above is mainly achieved as follows: 1) Cropping or adjusting the size of the input image to reduce the computational cost of the image segmentation. However, this approach realizes the loss of spatial information \cite{1Wu2017RealtimeSI,2Zhao_2018_ECCV}. 2) Increasing the speed of model inference by reducing the number of channels for semantic segmentation, which successively reduces the space capacity of the model \cite{37803544,4Chollet_2017_CVPR,5Paszke2016ENetAD}. 3) In pursuit of a compact framework, part of the down-sample layers may be abandoned, which reduces the receptive field of the model and become insufficient to cover large objects. Notably, this approach may be associated with poor discrimination ability\cite{5Paszke2016ENetAD}.

Researchers developed a U-shape network structure to compensate for the loss of spatial details, which gradually improves spatial information \cite{37803544,5Paszke2016ENetAD,6Xie_2015_ICCV,26wang2021pai}. The missing details are filled by fusing the hierarchical features of the backbone network. However, this method has two disadvantages: 1) The complete U-shaped structure extends the calculation of the model as it introduces high-resolution feature maps for additional calculations. 2) Challenges with restoring the spatial information cropped in the semantic segmentation model through simple upsampling and fusion. Therefore, the U-shaped structure is not the optimal solution, and we need to find a more lightweight and efficient structure.

Real-time semantic segmentation tasks have high demands for rich spatial information and multiscale semantic information. BiSeNet \cite{7Yu_2018_ECCV} employs a two-stream structure to replace the U-shaped structure and encodes spatial features and semantic information separately to produce excellent segmentation effects. However, the independent semantic encoding branch of BiseNet generates time-consuming calculations. Furthermore, pre-trained models from other tasks (including image classification) in the semantic branch of BiseNet are inefficient in semantic segmentation tasks. In STDC-Seg \cite{8Fan_2021_CVPR} (Short-Term Dense Concatenate Segmentation) network, a lightweight STDC backbone is designed to extract features. It eliminates feature redundancy on branches and utilizes edge detail information from ground truth to guide the spatial features learning. The STDC-Seg network has achieved satisfactory results in accuracy and speed; however, it does not consider the effect of different scale images on the network. A previous study found different segmentation results of images with different scales in the same network\cite{9Tao2020HierarchicalMA}.  The segmentation accuracy of small objects is low in small-scale images but can achieve excellent results in large-scale images. On the other hand, the segmentation effect of large objects (especially background) is poor in large-scale images but can be distinguished well in small-scale images. Therefore, we integrate the hierarchical multiscale attention mechanism into the STDC-Seg network to allow the model to learn the relationship of regions between different scales through attention. The model calculates the images of different scales combined with multiscale attention and learns high-quality features of different scales. Simultaneously, STDC-Seg does not consider the problem of feature alignment during feature aggregation in the ARM module. Direct relationship between the pixels of the local feature map and the upsampled feature map causes inconsistency of the context, further decreasing the classification accuracy in the prediction. To solve this problem, we integrated a feature alignment module (FAM)\cite{21Huang2021FaPNFP} in the STDC-Seg network.

Our STDC-MA network is based on the STDC2 backbone network of the STDC- Seg. STDC-MA integrates hierarchical multiscale attention \cite{9Tao2020HierarchicalMA} into the STDC-Seg. We integrated the attention regions of images at different scales of one image into the segmentation results of the STDC-MA network. This approach improves the effective application of multiscale features and solves the problem of rough segmentation in some regions, achieved using a single-scale image.

At the same time, we employed the feature alignment module (FAM) and feature selection module (FSM) described previously \cite{21Huang2021FaPNFP} to replace the original ARM module. This strategy (i)solves the problem of pixel offset associated with high-level feature upsampling and (ii) realizes the effective combination of high-level features and low-level features. As a result,  the segmentation results become more accurate on small objects. We tested the accuracy of our model using a verification dataset of Cityscapes.\cite{20Cordts_2016_CVPR} Under the input of 0.5x scale, the segmentation result of STDC-MA reached 76.81\% mean intersection over union (mIOU), 3.61\% higher than STDC-Seg.

\section{Relate Works}

\subsection{Lightweight Network}
The segmentation network relies on a robust feature extraction network as the backbone network to obtain sufficient semantic information. The backbone network mainly extracts the main features in the image, and its structure significantly impacts the performance of the segmentation network. ResNet \cite{10He_2016_CVPR}, for instance, utilizes a residual block to achieve excellent feature extraction effects.

The backbone network is the most computationally intensive part of the entire semantic segmentation model. Researchers have shifted their focus to the lightweight design of the backbone network to pursue real-time segmentation speed. MobileNetV1\cite{11Howard2017MobileNetsEC} uses deep separable con-volution to reduce FLOPs (Floating-point operations, used to measure the complexity of algorithms/models) in the inference stage. SqueezeNet \cite{12Iandola2016SqueezeNetAA} employs several $1\times1$ convolutions to replace $3\times3$ convolutions and uses the Fire module to reduce the model parameters. ShuffleNetV1\cite{13Zhang_2018_CVPR} proposes the channel shuffle operation to compensate for the lack of information exchange between point-by-point grouping convolutions and reduce network calculations. GhostNet \cite{14Han_2020_CVPR} adopts a few primitive convolution operations plus a series of simple linear changes to generate more features to reduce the overall parameters and calculations. The lightweight network designs demonstrate excellent performance in semantic segmentation. However, originally, the purpose of the network designs was to achieve image classification. Of note, a few adjustments or module improvements are warranted when applying the backbones to semantic segmentation.

\subsection{Real-time Semantic Segmentation}
The two strategies to ensure segmentation accuracy and speed in real-time semantic segmentation include 1) Lightweight backbone network. LRNet\cite{159106038} adopts factorized convolution block (FCB) to establish long-distance relationships and implement a lightweight and efficient feature extraction network. DFANet \cite{16Li_2019_CVPR} uses a lightweight network to reduce computational costs and develop cross-level aggregation modules for improving segmentation accuracy. 2) Multi-branch structure BiseNetV1 \cite{7Yu_2018_ECCV} proposes a two-stream structure of low-level details and high-level background information. STDC-Seg \cite{8Fan_2021_CVPR} inherits the BiseNet two-stream structure, optimizes the extractor of low-level and high-level details to achieve better performance. However, these real-time semantic segmentation models fail to consider the impact of inputs of different scales on the model. Here, we designed the STDC-MA network based on the work of STDC-Seg to compensate for the image scale impact.

\subsection{Muitiscale context methods}
The backbone network with a low output stride can solve better the fine detail problem in image segmentation. However, this design reduces the receptive field, posing difficulty to the network to predict large objects in the scene. Notably, the pyramid structure can attenuate the impact of the receptive field reduction by cascading the multiscale contexts. In \cite{27wang2021bilateral} high-level features of the encoder structure integrate all channel maps through dense channel relationships learned by the channel correlation coefficient attention module to refine the mask of output.  PSPNet \cite{17Zhao_2017_CVPR} utilizes a spatial pyramid cascade module. The combination of features  in the last layer of the module with multiscale features is achieved through a series of convolution operations. DeepLab \cite{187913730,19Chen2017RethinkingAC} employs atrous spatial pyramid pooling (ASPP) with different dilations of convolution to create denser features. Hierarchical multiscale attention mechanism \cite{9Tao2020HierarchicalMA} realizes dense feature aggregation between any two scales by learning the attention relationship between images of different scales. We integrated the hierarchical multiscale attention mechanism based on the STDC-Seg work to solve the impact of different scales on the segmentation work. In \cite{28jiao2021salient}, they proposed a regional growth algorithm based on the Gaussian pyramid to refine the edge of the output mask.

\subsection{Feature Alignment}
The Feature Alignment Module aligns the semantic relationship between different feature maps in the feature fusion module. It ensures that the feature relationship of the context does not produce large deviations, which successively improves model segmentation accuracy. In the SegNet \cite{37803544} network, the encoder stores the position information of the maximum pooling and employs the index of the maximum pooling in the decoder for upsampling. The RoI Align,\cite{22He_2017_ICCV} avoids quantization calculation, and the value of each RoI is calculated by bilinear interpolation, solving the problem of feature misalignment associated with quantization in RoI Pooling. EDVR \cite{23Wang_2019_CVPR_Workshops} utilizes the PCD Alignment module constructed via deformable convolution \cite{24Dai_2017_ICCV} to achieve feature alignment on a single scale.

\begin{figure*}[!t]
   \centering
   \includegraphics[width=1\textwidth]{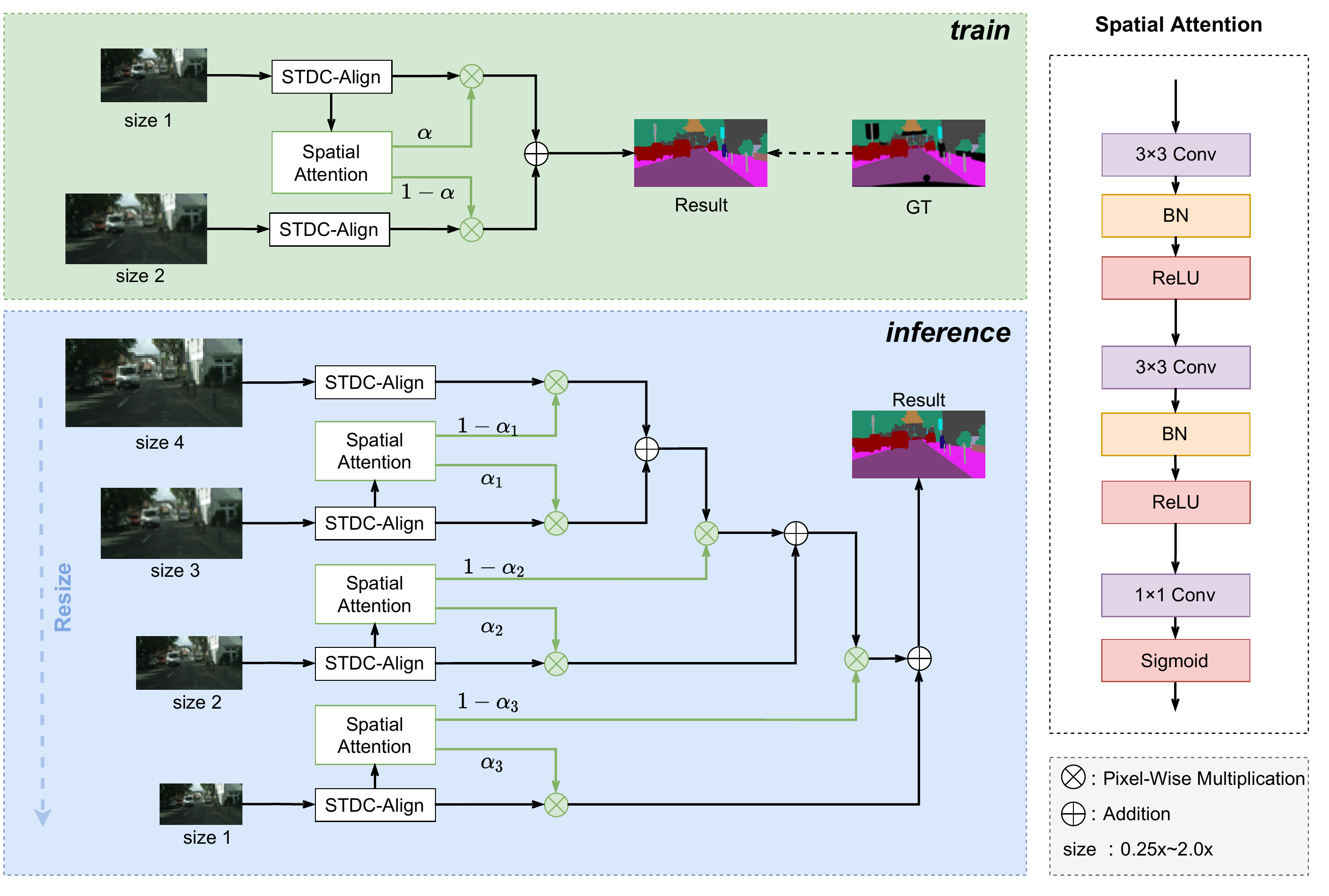}
   \caption{The structure of the STDC-MA network. The STDC-Algin denotes the STDC-Seg network with the feature alignment module (FAM) added. The Spatial Attention Module denotes the hierarchical multiscale attention module. All Spatial Attention modules in the inference use the same parameters.}
   \label{total}
\end{figure*}

\section{Proposed Methods}
\subsection{Short-Term Dense Concatenate with Multiscale Attention and Alignment Network}
Our work employs the feature alignment module \cite{21Huang2021FaPNFP} and the hierarchical multiscale attention mechanism \cite{9Tao2020HierarchicalMA} to the STDC-Seg network and designs a short-term dense concatenate with multiscale attention and alignment (STDC-MA) network. The Feature Alignment Module  learns the offset between high-level and low-level features and introduces a feature selection module to generate low-level feature maps with rich spatial information. This method combines the offset with enhanced low-level features. It solves the problem of pixel offset during the fusion of high-level and low-level features, fully utilizing the high-level and low-level image features. The hierarchical multiscale attention mechanism learns the relationship of attention regions from two different input sizes of one image to compound the attention from different receptive fields. This approach reduces the unfocused regions of the input image, fully utilizing the multiscale features to solve the problem of rough mask edges. The STDC-MA network structure is shown in Fig. \ref{total}.


The network learns hierarchical multiscale attention between the two scales of 1.0x and 0.5x at once during training. In inference, hierarchical multiscale attention fusion is used according to the number of input images with different scales. Numerous hierarchical multiscale attention modules are shown in Fig. \ref{total}. In practice, a similar hierarchical multiscale attention module uses the same parameters. Compared to the separated attention for different scales, this design significantly reduces the parameters. Our STDC-MA network achieves 76.81\% mIOU on Cityscapes \cite{20Cordts_2016_CVPR} validation data set with input image (scale, 0.5), 3.61\% higher than STDC-Seg\cite{8Fan_2021_CVPR}.

\subsection{Hierarchical Multiscale Attention}
Investigation of hierarchical multiscale attention demonstrated that the output masks for different scale inputs differ even if the input is derived from the same image \cite{9Tao2020HierarchicalMA}.  Images of different scales contain different spatial information. For instance, large-scale images have detailed spatial information, and extraction of semantic features is challenging. Therefore, small objects are segmented accurately in the segmentation results of high-scale input images, while large objects have rough segmentation. On the other hand, the spatial information of the low-scale image is rough, and the semantic feature is easy to extract. Therefore, large objects are segmented accurately in the segmentation results of low-scale input images, while small objects have rough segmentation.

Taking full advantage of different scales to refine the output of the segmentation network is problematic. As such, hierarchical multiscale attention proposes to learn the relationship between the attention regions at different scales of one image to integrate the attention regions in different receptive fields. This method reduces the unfocused areas of the input image and improves the segmentation accuracy of the network for small target objects.

\begin{figure}
   \centering
   \includegraphics[width=1\textwidth]{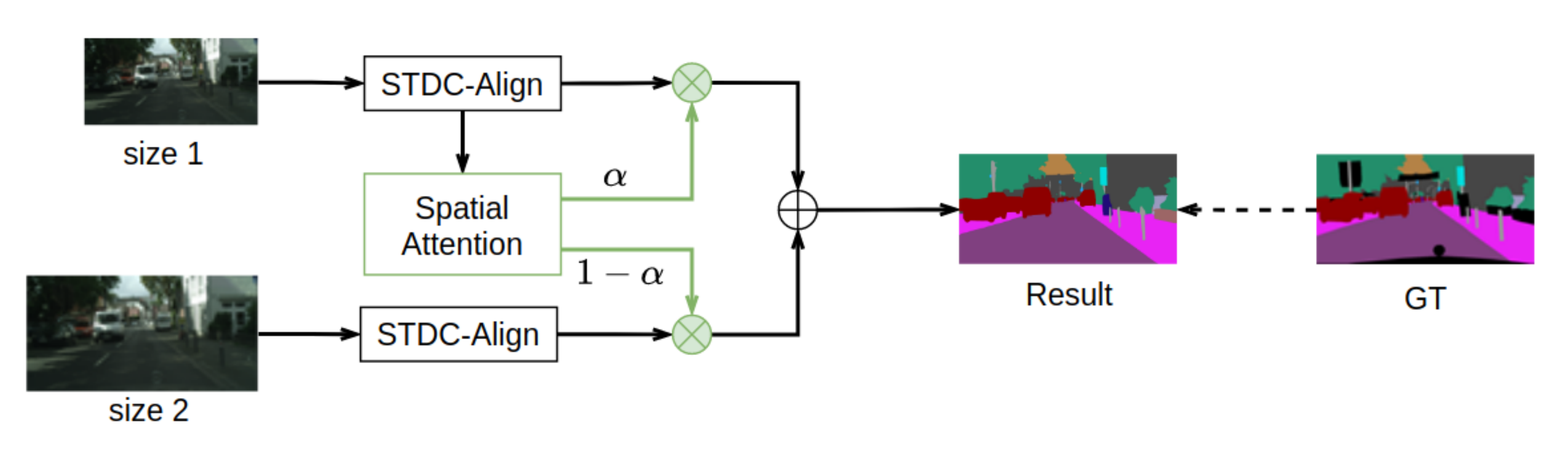}
   \caption{Multiscale fusion of any two-scale of inputs. The structure of the Spatial Attention module is shown in Fig. ~\ref{total}.}
   \label{p2}
\end{figure}

\begin{equation}
   \label{euqation1}
   P_{result} = G(S_i) \times \alpha_i + G(S_{i+1}) \times (1-\alpha_i)
\end{equation}

\begin{figure*}
   \centering
   \includegraphics[width=1\textwidth]{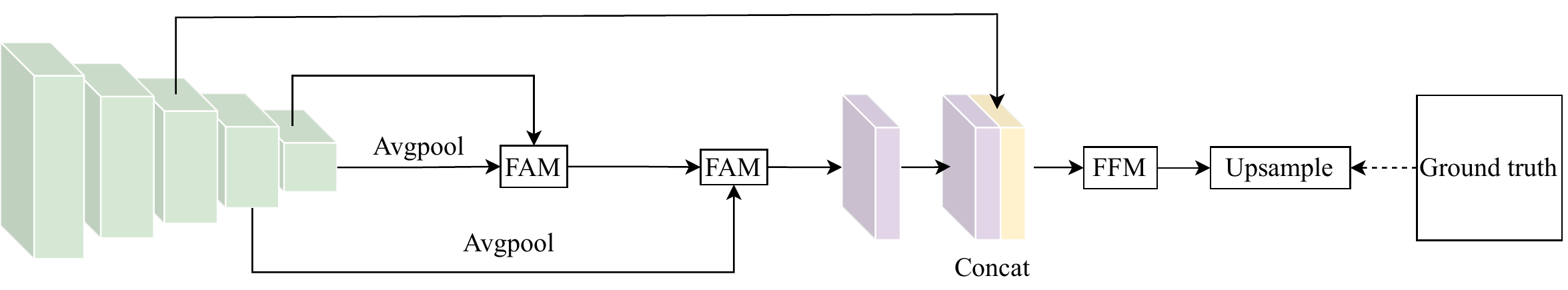}
   \caption{The structure of the STDC-Align network. FAM denotes the Feature Alignment Model. FFM denotes the Feature Fusion Module in STDC-Seg\cite{8Fan_2021_CVPR}.}
   \label{p3}
\end{figure*}

The ASPP in DeepLab \cite{187913730,19Chen2017RethinkingAC} utilizes dilation convolutions to create a denser feature aggregation. Although a larger receptive field was obtained in these designs, different areas of interest corresponding to different scales were not recognized clearly.
Hierarchical multiscale attention differs from previous attention mechanisms, which focus on single feature maps. Hierarchical multiscale attention learns the relationship between any two input scales, effectively reducing the consumption of excessive attention mechanism calculations.

Let $S = {S_1, S_2, . . . , S_N }$ denotes the collection of images with different N scales. $S_i(1 \leq i \leq N )$ denotes the $i^{th}$ scale of the image, and the scale of $S_i$ is smaller than the $S_{i+1}$. The fusion of hierarchical multiscale attention modules involves a series of fusions between any higher-level feature map and the corresponding lower-level feature map (Fig \ref{p2}). The feature fusion of $S_i$ and $S_{i+1}$ is defined as:

Where $P_{result}$ denotes the output of the inputs in $S_i$ and $S_{i+1}$. $G(\cdot)$ denotes a segmentation network; $\alpha_i$ denotes the hierarchical multiscale attention between $S_i$ and $S_{i+1}$.

The hierarchical multiscale attention is integrated into the STDC-Align network to determine the feature relationship between different scales, guiding the extraction of different regions of interest to refine the segmentation mask. Here, we propose the final semantic segmentation model STDC-MA, which improves the segmentation accuracy of small objects.

\subsection{Short-Term Dense Concatenate Align Network}
The short-term dense concatenate network (STDC-Seg) \cite{8Fan_2021_CVPR} follows the two-stream design structure of BiseNetV1 \cite{7Yu_2018_ECCV}. It employs the STDC as the backbone to extract both semantic and spatial features, establishing an efficient and lightweight design. The ARM module of STDC-Seg is a feature aggregation module that does not consider the problem of pixel offset during feature aggregation between different feature maps, which is solved by a practical feature alignment module. In SegNet \cite{37803544}, the encoder employs the position of maximum pooling to enhance upsampling. Of note, the problem of pixel shift is solved but part of the feature information is lost in the image after max pooling, which cannot be compensated for by upsampling.

In Feature Alignment Module (FAM) \cite{21Huang2021FaPNFP}, the feature selection module (FSM) is applied to enhance the rich spatial information of low-level feature maps, ensuring that the final alignment result is as close to the ground truth as possible. To solve the problem of pixel misalignment, our  method employs deformable convolution (DCN)\cite{24Dai_2017_ICCV} to learn the feature offset between two feature maps. Then the model uses the offset to guide the procedure of feature alignment. The FAM module achieves the same effect for feature map fusion as the ARM aggregation module in the STDC-Seg network. Also, the parameters of the FAM module are 1.3M lower than the ARM module. In this manner, we replaced the ARM aggregation module with the Feature Alignment Module (FAM) and proposed an STDC-Align network, structured as demonstrated in Fig. \ref{p3}.

\begin{figure}[!h]
   \centering
   \includegraphics[width=1\textwidth]{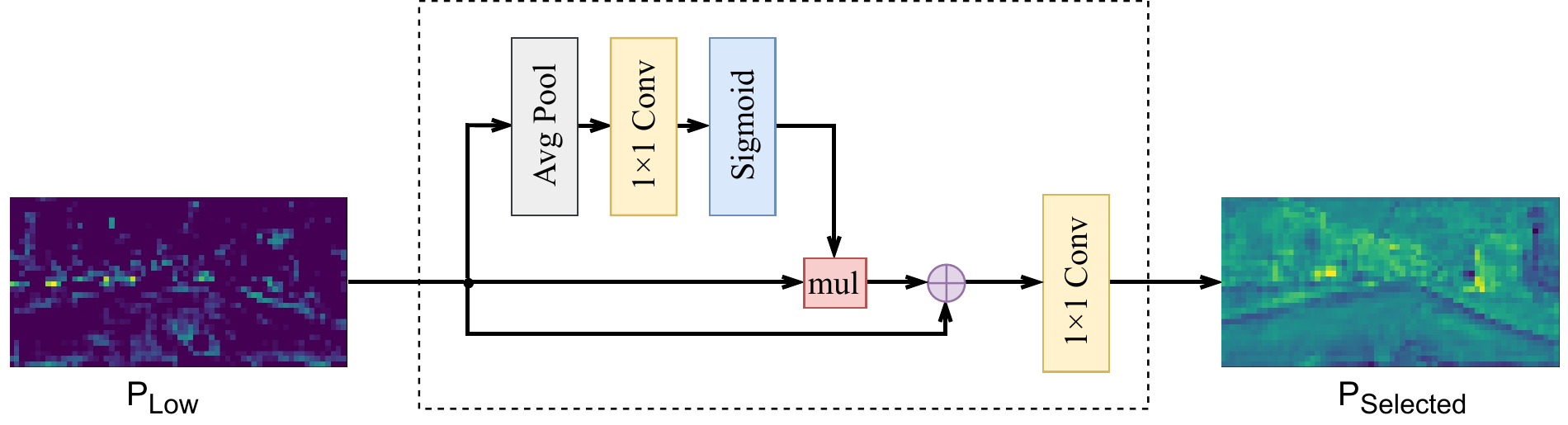}
   \caption{The structure of the feature selection module. The upper branch denotes channel attention. Mul denotes multiplication.\label{FSM}}
\end{figure}

\subsection{Feature Alignment and Feature Selection Module}
\subsubsection{Feature Selection Module}
The feature selection module (FSM) utilizes channel attention (corresponding to the upper branch of Fig. \ref{FSM}) to enhance the spatial information in the low-level features. This process is defined as:

\begin{equation}
   \label{equation2}
   P_{selected} = \phi(P_{low})
\end{equation}

\begin{equation}
   \label{equation3}
   \phi(P_{low}) = Conv(\sigma(W_{selection}P_{low})\times P_{low} + P_{low})
\end{equation}

Where $P_{selected}$denotes the feature map after feature selection; $P_{low}$ denotes the low-level feature map; $\phi(\cdot)$ denotes the feature selection process corresponding to the FSM, which successively selects the features of the current feature map; $Conv$ denotes $1 \times 1$ convolution; $\sigma(\cdot)$ denotes the sigmoid function; $W_{selection}$  denotes learnable parameters. In the implementation, the learned parameters, $W_{selection}$and $P_{low}$, are constructed into channel attention to realize the selection function of the feature selection module. The structure of the feature selection module is outlined in Fig. \ref{FSM}.

\subsubsection{Feature Alignment Module}
Feature alignment module (FAM) employs deformable convolution (DCN) \cite{24Dai_2017_ICCV} to learn the offset between the high-level feature map and the FSM-derived feature map. This method utilizes the offset to achieve feature alignment and fusion between $P_{selected}$ and high-level feature map $P_{high}$. The aligned feature map is denoted by $P_{aligned}$. This process is defined as:

\begin{equation}
   \label{equation4}
   P_{aligned} = \psi(P_{selected},P_{high})
\end{equation}

\begin{equation}
   \label{equation5}
   P_{aligned} = f([Conv([P_{selected},P_{high}]),P_{high}])+P_{selected}
\end{equation}

\begin{figure}
   \centering
   \includegraphics[width=1\textwidth]{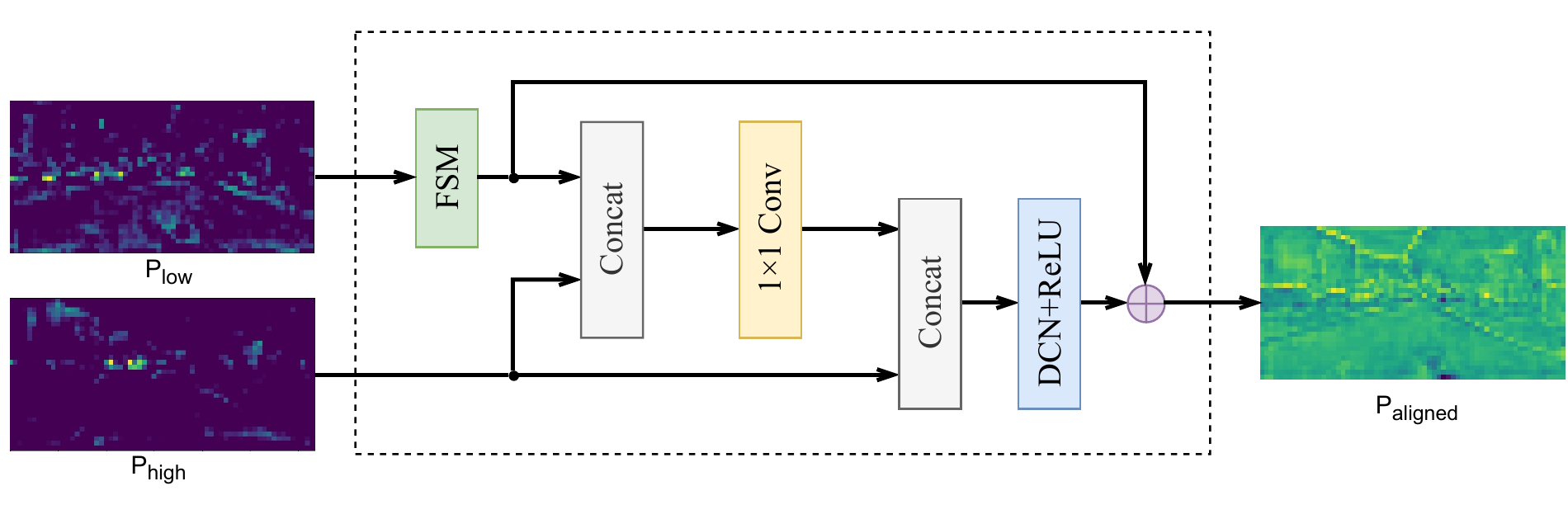}
   \caption{Structure of the feature alignment module. DCN denotes the deformable convolution.\label{FAM}}
\end{figure}

Where $P_{aligned}$ denotes the aligned feature map; $f(\cdot)$ denotes the deformable convolution (corresponding to DCN in Fig. \ref{FAM})); $Conv$ denotes $1 \times 1$ convolution; $[\cdot,\cdot]$ denotes the channel-wise concat of two feature maps.

In implementing the feature alignment module, the high-level feature map is upsampled to the same size as the feature map selected by the feature selection module before concatenating. At the same time, the deformable convolution is employed to calculate the concatenate result to achieve the effect of feature alignment. Lastly, the selected feature map and the aligned feature map are added by pixel. The structure of the feature alignment module is shown in Fig. \ref{FAM}.

\section{Experimental Results and Discussion}
\subsection{Dataset}
The presently established method is implemented using the Cityscapes \cite{20Cordts_2016_CVPR} dataset, a widely used semantic scene analysis dataset, including scenes between different cities from the perspective of a vehicle-mounted camera. Cityscapes contain 5000 fine annotated images divided into training, validation, and test sets with 2975, 500, and 1525 images, respectively. The dataset comprises 30 classes of labels, 19 of which are utilized for semantic segmentation tasks; these images have a high resolution of 2048x1024. In most cases, the Cityscapes data set is used in pre-training models of vision models for autonomous driving, therefore, poses a challenge for semantic segmentation tasks. In our investigation,  we used the fine annotated training set and evaluated our model on the validation set.

\subsection{Training Details and Evaluation Indicators}
The model developed herein was trained on a Tesla A100 with image input resolution of 1024x512 and Adam as the optimizer. The initial learning rate was set to 0.0001, $\beta = (0.9,0.999)$, $\epsilon = 1e-08$, no weight decay. A total of 60,000 iterations were trained using batch size 8, and mIOU was applied for validation, measuring the overlap degree between segmentation results and ground truth.

\subsection{Ablation Study}
In this section, the effectiveness of each part of the STDC-MA network was verified gradually. However, in future experiments, we shall make improvements based on the work of the STDC-Seg \cite{8Fan_2021_CVPR} network and evaluate the model on the Cityscapes \cite{20Cordts_2016_CVPR} validation data set.

\subsubsection{Ablation for Feature Alignment Module}
The present research found that the ARM module in the STDC-Seg network \cite{8Fan_2021_CVPR} is a feature aggregation module between different feature maps. Notably, because this module does not account for feature alignment, it was substituted with the feature alignment module (FAM)\cite{21Huang2021FaPNFP}, and we proposed our STDC-Align network. The analysis demonstrated that, at the input scale of 0.5, our STDC-Align network achieved 73.57\% mIOU, 0.37\% higher than STDC-Seg. Furthermore, the parameters of our STDC-Align network were 21.0M, 1.3M less than that of STDC-Seg.

\subsubsection{Ablation for Hierarchical Multi-scale Attention}
Here, the hierarchical multiscale attention mechanism \cite{9Tao2020HierarchicalMA} is utilized in the STDC-Seg network \cite{8Fan_2021_CVPR}, with the view that this method can identify different parts of interest between different scales and achieve complementary advantages. Scale images (0.5x and 1.0x) are used as input for training to learn the attention relationship between two different scales. Subsequently, the results of different scale combinations (the scale can be chosen in [0.25x, 0.5x, 1.0x, 1.5x, 2.0x]) are tested on the Cityscapes \cite{20Cordts_2016_CVPR} validation data set. The results are presented in Table \ref{table1}.

\begin{table*}[ht]
   \caption{The performance of different scale combinations. Scale combination denotes the input combinations in a hierarchical multiscale attention mechanism.}
   \label{table1}
   \begin{center}
      \resizebox{1\textwidth}{!}{
         \begin{tabular}{ccccccc}
            \hline
            \rule[-1ex]{0pt}{3.5ex} \textbf{Scale combination}     & 0.5x+1.0x & 1.0x+2.0x & 0.5x+1.0x+1.5x & 1.0x+1.5x+2.0x & 0.5x+1.0x+1.5x+2.0x & 0.25x+0.5x+1.0x+1.5x+2.0x \\
            \hline
            \rule[-1ex]{0pt}{3.5ex} \textbf{\textbf{Mean IOU(\%)}} & 72.98     & 74.55     & 75.68          & 76.11          & 76.53               & \textbf{76.55}            \\
            \hline
         \end{tabular}}
   \end{center}
\end{table*}

\subsection{Comparison of our methods and output mask with STDC-Seg}
The segmentation result of the STDC-MA network achieved higher performance in mIOU, providing evidence that our method is effective. Table \ref{table2} shows the performance indicators of our network. Compared to the structure of the STDC-Seg network\cite{8Fan_2021_CVPR}, the structure of the STDC-MA network adds a hierarchical multiscale attention mechanism. It employs a feature alignment module to replace the ARM module, decreasing 0.1M parameters and increasing 3.61\% mIOU.

\begin{table}[ht]

   \caption{Performance of STDC-Seg network and our method. STDC-Seg is the baseline of our work. STDC-Align network replaces the ARM module with the FAM module. STDC-MA represents the final network proposed in this article.
      \label{table2}}
   \begin{center}
      \resizebox{0.5\textwidth}{!}{
         \begin{tabular}{cccc}
            \hline

            \rule[-1ex]{0pt}{3.5ex} \textbf{Method}                    & \textbf{Flops}  & \textbf{Parmas} & \textbf{Mean IOU(\%)} \\
            \hline
            \rule[-1ex]{0pt}{3.5ex} STDC-Seg\cite{8Fan_2021_CVPR}      & 73.32G          & 22.3M           & 73.20                 \\
            \rule[-1ex]{0pt}{3.5ex} STDC-Align(our)                    & \textbf{72.14G} & \textbf{21.0M}  & 73.57                 \\
            \rule[-1ex]{0pt}{3.5ex} STDC-Seg+Muti-scale Attention(our) & 103.73G         & 23.4M           & 76.55                 \\
            \rule[-1ex]{0pt}{3.5ex} STDC-MA(our)                       & 102.27G         & 22.2M           & \textbf{76.81}        \\
            \hline
         \end{tabular}
      }
   \end{center}
\end{table}

\begin{figure*}
   \centering
   \includegraphics[width=1\textwidth]{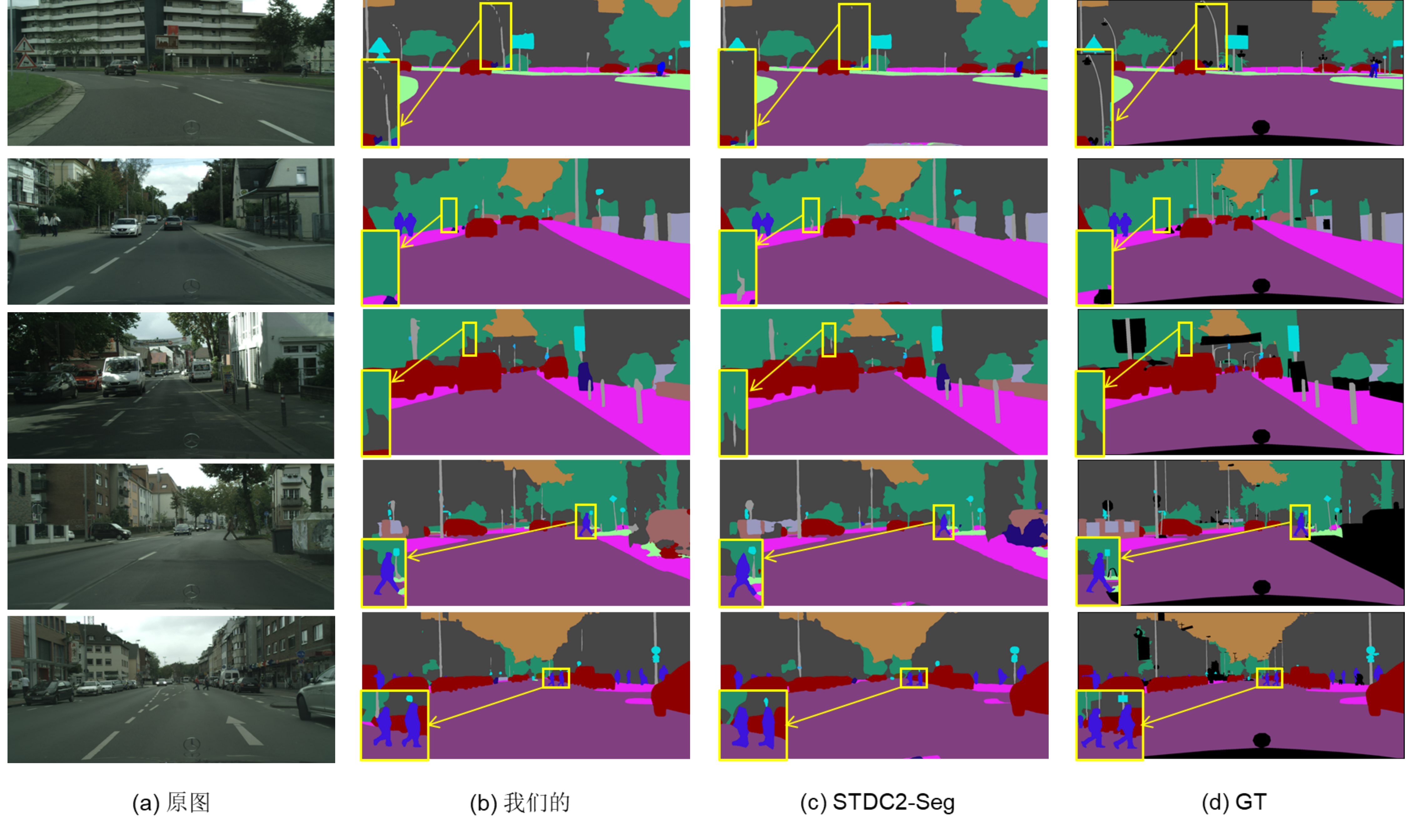}
   \caption{Comparison of the segmentation effect obtained by our approach and that of STDC-Seg network. (a) denotes the original image of the input segmentation network; (b) denotes the output result of the image through the STDC-MA network; (c) denotes the output result of the image through the STDC-Seg network; (d) denotes the ground truth.
      \label{result_stdc-ma}}
\end{figure*}

The output of the STDC-MA network  is shown in Fig. \ref{result_stdc-ma}. The presently developed method is smoother and more accurate on small objects. In the first row, our STDC-MA network  obtained a more accurate mask of street lights than the STDC-Seg network  \cite{8Fan_2021_CVPR}.  In the second and third rows, the STDC-Seg network mistakenly predicted the railings. In the fourth and fifth rows, our STDC-MA network demonstrates a smoother result in predicting the pedestrian, similar to the ground truth, and better than the STDC-Seg network.

\section{Conclusions}
The STDC-MA network, integrating hierarchical multiscale attention mechanism \cite{9Tao2020HierarchicalMA} and feature selection module \cite{21Huang2021FaPNFP}, is proposed for the semantic segmentation task. The hierarchical multiscale attention mechanism is employed to learn the relationship of attention regions from two different input sizes of one image. Through this relationship, the different regions that the attention is focused on are integrated into the segmentation results. The present method makes up for the defect of the STDC-Seg network in the multiscale concern problem and improves the accuracy of the small object segmentation.

\bibliographystyle{unsrt}
\bibliography{references}  

\end{document}